# A RobustICA-Based Algorithm for Blind Separation of Convolutive Mixtures

*Zaid Albataineh and Fathi M. Salem*
*Circuits, Systems, And Neural Networks (CSANN) Laboratory*
Department of Electrical and Computer Engineering, Michigan State University,
East Lansing, Michigan 48824-1226, U.S.A.
Email: albatain@msu.edu, salemf@msu.edu

***Abstract— We propose a frequency-domain method based on robust independent component analysis (RICA) to address the multichannel Blind Source Separation (BSS) problem of convolutive speech mixtures in highly reverberant environments. We impose regularization processes to tackle the ill-conditioning problem of the covariance matrix and to mitigate the performance degradation in the frequency domain. We apply an algorithm to separate the source signals in adverse conditions, i.e. high reverberation conditions when short observation signals are available. Furthermore, we study the impact of several parameters on the performance of separation, e.g. overlapping ratio and window type of the frequency domain method. We also compare different techniques to solve the frequency-domain permutation ambiguity. Through simulations and real-world experiments, we verify the superiority of the presented convolutive algorithm among other BSS algorithms, including recursive regularized ICA (RR-ICA), independent vector analysis (IVA).***

***Index Terms— Blind Source Separation (BSS), Independent Component Analysis (ICA), Robust Independent Component Analysis (RobustICA), highly reverberant environments, gradient descent algorithms, Recursive Regularized ICA (RR-ICA), Independent Vector Analysis (IVA).***

## I. INTRODUCTION

Blind Source Separation (BSS) [1] has a solid theoretical foundation and many potential applications. In fact, BSS has remained a very important topic of research and development for a long time in many areas, such as biomedical engineering, image processing, communication systems, speech enhancement, remote sensing, etc. BSS techniques do not require any prior knowledge about a mixing matrix or source signals and do not require any training data.

Independent Component Analysis (ICA) is a powerful tool in BSS and Multichannel Blind Deconvolution (MBD) [1], [2]. ICA is a key factor of BSS and unsupervised learning algorithms. ICA techniques do not assume full *a priori* knowledge about the mixing environment, source signals, etc. Also, ICA is related to Principle Component Analysis (PCA) and Factor Analysis (FA) in multivariate analysis and data mining. This is especially the case when corresponding to second order methods in which the components or factors are in the form of a Gaussian distribution [1], [2], [6]. Nevertheless, ICA is a statistical technique that includes higher order statistics (HOS), where the goal is to represent a set of random variables as a linear transformation of statistically independent components [1].

ICA methods usually assume certain properties of the sources or mixing system in order to exploit a separation criterion which imposes the same properties on their estimates. In ICA of speech signals, several approaches have been proposed in a simple case of instantaneous linear mixtures [6-15]. However, convolutive linear mixtures are considered more suitable in real-world applications [1-3]. Several convolutive ICA approaches have been proposed for time domain [3], [4], [42-43] and frequency domain [32-40] methods. Refer to [3], [22] for more details of existing convolutive ICA methods.

One can exploit the inherent non-stationary attribute of natural speech signals by using the second order statistics (SOS) method [2]. Mixing environments are considered to be stationary environments and, even in a short period, one can exploit the higher order statistics e.g. Joint Approximation Diagonalization (JAD) problem as in [9], [10]. According to [42], [46], online BSS algorithms can be adapted in the time domain under non-stationary conditions. However, the time domain approach suffers from slow convergence, lack of stability and high computational complexity [22].

Alternatively, a block online frequency domain BSS algorithm is proposed in [22], [28]. In this case, one can apply the separation processes on individual blocks of the input data over time. Furthermore, one can assume that the mixing environment is stationary in short time windows. This means that the source signals don't change their location during this interval of time. This requires choosing the right time frame to guarantee that the separation algorithms are accurate enough with this given observed data within this window. For more details, refer to [43], in which there is a recent ICA algorithm based on the time domain framework for the short mixtures.

The proposed recursive regularized ICA [40] algorithms allow estimating a large number of demixing matrices even with a short amount of data. Despite the good performance of the aforementioned algorithms, it is considered to be semi-blind since it is based on prior knowledge about the acoustic source signals, i.e.: the acoustic propagation and the spectral characteristic of the source signals. In [22], [31], Authors studied the relationship between the number of frames of the STFT analysis and the BSS algorithms based on frequency framework. They argued the BSS algorithms in frequency domain are significantly affected by the number of the mixing matrices. Also in [22], [31], Authors proposed a method of applying the ICA adaptation to a group of frequencies in order to leave the size of the STFT large enough to achieve accurate separation processes. This method assumed that the acoustic propagation approximated is based on an anechoic model, i.e. as the DRR decreases.

There are several drawbacks for separating the acoustic sources based on frequency domain methods [22], [40]. First of all, when we have a high reverberation environment, this requires us to increase the number of demixing matrices to ensure an efficient estimation for the source signals. However, this requirement is not easy to satisfy, especially if we have short observation signals of the source signals. Therefore, inspired by the works of V. Zarzoso,P. Comon [5], [6], this paper considers several challenges for the convolutive mixtures in the frequency domain in order to carry out the RobustICA-based algorithm in the frequency domain. We can summarize these challenges as follows.

- *Increasing the immunity of the BSS algorithm towards the outlets, e.g. signals' length, additive noise, reverberation time and source moving etc.*
- *Implementing should be optimized to be suitable for the real-time operation [42] in order to make the real-time DSP processor handle the computational cost without interruptions or distortions.*
- *Effectively treating the scaling and permutation problems in the frequency domain.*
- *Reducing the computational complexity of the ICA algorithms based on the frequency framework.*
- *Controlling the accuracy of the ICA algorithm especially when short mixtures are available and the demixing matrices are not constrained by any anechoic model.*

Regarding the paper's notation of matrix computation, a matrix is denoted as a bold capital letter such as $\mathbf{A}$, $\mathbf{A}^{\mathrm{T}}$ is the matrix transpose of $\mathbf{A}$, its Frobenius norm is marked by $\|\mathbf{A}\|$, an identity matrix of size n is denoted as $\mathbf{I}_{\mathrm{n}}$, a vector is denoted as bold small letter such as $\mathbf{a}$, and scalars are denoted as a small letter such as a.

The remainder of the paper is organized as follows. Section II provides a brief description of convolutive mixtures and the problem statement. Section III presents the RobustICA-based method in the frequency domain. In Section IV, we perform solving the ambiguities in the ICA algorithm based on the frequency domain. The comparative experiments' results and conclusions are given in Section V and Section VI, respectively.

## II. Convolutive Mixtures

A convolutive mixture can be considered a natural extension of the instantaneous BSS problem. Assume an $M$-dimensional vector of received discrete time signals $\mathbf{x}(\mathrm{k}) = [\mathbf{x}_1(\mathrm{k}), \mathbf{x}_2(\mathrm{k}), \dots, \mathbf{x}_{\mathrm{M}}(\mathrm{k})]^{\mathrm{T}}$ is to be produced at time k from an $N$-dimensional vector of source signals $\mathbf{s}(\mathrm{k}) = [\mathbf{s}_1(\mathrm{k}), \mathbf{s}_2(\mathrm{k}), \dots, \mathbf{s}_{\mathrm{N}}(\mathrm{k})]^{\mathrm{T}}$, where $\mathrm{M} \geq \mathrm{N}$, by using a stable mixture model [2]:

$$\mathbf{x}(\mathrm{k}) = \sum_{\mathrm{p}=-\infty}^{\infty} \mathbf{H}_{\mathrm{p}}\mathbf{s}(\mathrm{k}-\mathrm{p}) = \mathbf{H}_{\mathrm{p}} * \mathbf{s}(\mathrm{k}),$$

$$\text{with } \sum_{-\infty}^{\infty} \|\mathbf{H}_{\mathrm{p}}\| \leq \infty \qquad (1)$$

where $*$ represents the linear convolution operator and $\mathbf{H}_{\mathrm{p}}$is an (M x N) matrix of mixing coefficients at time-lag p.

### *A. Problem Definition*

Assume that elements $\mathrm{h}_{\mathrm{jip}}$ denote the coefficients of the Finite Impulse Response (FIR) filter $\mathbf{H}_{\mathrm{p}}$ and L is the maximum unknown channel length. Then, the noise-free convolutive model is written as follows:

$$\mathbf{x}(\mathrm{k}) = \sum_{\mathrm{p}=0}^{\mathrm{L}-1} \mathbf{H}_{\mathrm{p}}\mathbf{s}(\mathrm{k}-\mathrm{p}) \qquad (2)$$

Thus, one can find an approximate inverse channel matrix $\mathbf{W}_{\mathrm{p}}$ in order to recover the source signals $\mathbf{s}(\mathrm{k}) = [\mathbf{s}_1(\mathrm{k}), \mathbf{s}_2(\mathrm{k}), \dots, \mathbf{s}_{\mathrm{M}}(\mathrm{k})]^{\mathrm{T}}$ such that

$$\mathbf{y}(\mathrm{k}) = \mathbf{W}_{\mathrm{q}} * \mathbf{x}(\mathrm{k}) = \sum_{\mathrm{q}=-0}^{\mathrm{Q}-1} \mathbf{W}_{\mathrm{q}}\mathbf{x}(\mathrm{k}-\mathrm{q}) = \hat{\mathbf{s}}(\mathrm{k}) \qquad (3)$$

where Q is the length of the inverse of the channel impulse response. There are two approaches to solve this problem and recover the source signals.

*Time domain approaches* have several general drawbacks; for example, Q should be selected at least equal to the unknown true channel L. Therefore, for a long mixing filter, which means long transfer functions, the computation will be too expensive [2], [14], [22]. Using the IIR filter instead of the long FIR filter to overcome this problem causes increased instability and might require inversion of the non-minimum phase filters [2], [3], [22]. Moreover, time approaches are sensitive to channel order mismatch [3]. That said, time domain methods are suitable and very efficient for small mixing filters such as in a communication channel [2], [12].

Because of all these limitations to time domain approaches, we focus our study on *frequency domain approaches* to solve the BSS problem. The main advantage of a frequency domain BSS approach is the ability to apply the set of any instantaneous ICA algorithms to solve the convolutive BSS problem. On the other hand, the main challenges of BSS in the frequency domain are permutation and scaling ambiguities. Refer to [1], [3], [22] for a recent survey. However, one can re-map the aforementioned BSS models into the frequency domain by applying the Discrete Fourier Transform (DFT) on the observed signals $\mathbf{x}(\mathrm{k})$ in order to transform it to the instantaneous mixtures problem as follows:

$$\mathbf{x}(\mathrm{k}) = \mathbf{H} * \mathbf{s}(\mathrm{t}) \leftrightarrow \mathbf{x}(\mathrm{q},\mathrm{w}) \cong \mathbf{H}(\mathrm{w})\mathbf{s}(\mathrm{q},\mathrm{w}) \qquad (4)$$

where w is a frequency index, q is a frame index, $\mathbf{s}(q,w)=[\mathbf{s}_1(q,w),\dots,\ \mathbf{s}_N(q,w)]^T$ and $\mathbf{x}(q,w)=[\mathbf{x}_1(q,w),\dots,\ \mathbf{x}_M(q,w)]^T$. The previous equation is considered to be valid only for periodic signals $\mathbf{s}(t)$. However, it is approximately valid if the time-convolution is circular. Therefore, ensuring that the time convolution is circular [1] requires making the Fourier Transform length significantly larger than the maximum length of the mixing channels L [6]. In [28], [40], researchers imposed the spectral smoothing approach in order to mitigate the circularity effect in frequency domain BSS methods. In practice, to avoid the convergence into local minima during the separation processes, one can separate the observed signal at each frequency bin. Thus, the sampled observed signals $\mathbf{x}_s(t)$ are sampled at the discrete time instant $n_s$ using the sampling frequency $f_s$. Then one can transform the sampled signals into time-frequency domain $\mathbf{x}(q,w)$ using the short time Fourier transform (STFT) applied to T overlapped samples of the observed signals. However, one can express the time-frequency of the *mth* sensor at *t* frame as follows

$$\mathbf{x}(q,w)=\sum_{n_s}\mathbf{x}_s(n_s)\mathrm{win}\left(\frac{n_s-q.\mathrm{shift}}{f_s}\right)e^{-j2\pi w\frac{n_s}{f_s}}$$

$$\forall\ w=\frac{t}{T}f_s\ ,q\in\ [0,\dots,T-1] \qquad (5)$$

Where $\mathrm{win}(\cdot)$ denotes the windowing function, here, we usually use the Hanning window since it is typically for acoustic signals. The Hanning window [17] is given by

$$\mathrm{win}(n_s)=\frac{1}{2}\left(1+\cos\left(\frac{2\pi n_s}{T}\right)\right) \qquad (6)$$

In a real-world scenario, we use the reverberation time $T_{60}$ to approximately define the length of the impulse response, since the impulse response functions $\mathbf{h}(t)$ are theoretically infinite. The reverberation time $T_{60}$ is the required time that reduces the energy of sounds into 60 dB where the sound signal becomes no longer active or “dies away”. Therefore, the convolutive ICA model can be approximated into a series of the instantaneous ICA model as follows:

$$\mathbf{x}(q,w)=\mathbf{H}(w)\mathbf{s}(q,w) \qquad (7)$$

Where w represents the frequency bin, q denotes the time domain frame, e.g. in a short time frequency transform, $\mathbf{x}(q,w)$ is a column vector of the observed signals in frequency domain, $\mathbf{s}(q,w)$ is a column vector of the original source signals and $\mathbf{H}(w)$ is an M x N mixing matrix in frequency domain.

For the sake of simplicity, let us assume that the number of source signals N is equal to the number of the observed signals M, i.e., N = M. Thus, by applying the ICA algorithm to the $\mathbf{x}(q,w)$ at each frequency bin, one can recover the estimated source signals as follows:

$$\mathbf{y}(q,w)=\mathbf{W}(w)\mathbf{x}(q,w) \qquad (8)$$

where $\mathbf{W}(w)$ is the demixing matrix at wth frequency bin. Also, due to the well-known symmetry property of the Fourier Transform, one can simply find demixing matrices ($\mathbf{W}(w)$) as a half of the frequency bins $w\in(0,\dots,\frac{T}{2})$, and then use the symmetry property to find the others.

## III. The presented method Based on the RobustICA framework

In this section, a new strategy is proposed, based on the RobustICA method of the kurtosis framework. Here, one needs to first recall the time-frequency representation of the observed vector equation (2),

$$\mathbf{x}(q,w)=\mathbf{H}(w)\mathbf{s}(q,w) \qquad (16)$$

The aim of this study is to estimate the demixing matrix $\mathbf{W}$ from the observed vector $\mathbf{x}$ under the assumption that the impulse response of all mixing filters is assumed constant during the recording. The estimated source vector is given as the following at each frequency bin:

$$\mathbf{y}(q,w)=\mathbf{W}(w)\mathbf{x}(q,w) \qquad (17)$$

### A. Step1: Preprocessing (Data Whitening)

In the preprocessing step, the demixing matrices $\mathbf{W}(w)$ are detected up to a unitary matrix $\mathbf{U}(w)$ using the second order statistic (SOS). This step was used to reduce the noise and to eliminate redundancy in the data at each frequency bin. The N x N covariance matrix ($\mathbf{R}$) of the noise-free observed signals can be expressed by

$$R(q,w)=E[x(q,w)x(q,w)^H] \quad \forall\ w=0,\dots,\frac{T}{2} \qquad (18)$$

By substituting $\mathbf{x}(q,w)$ in (16), one gets $\mathbf{R}$ as follows

$$\mathbf{R}(q,w)=\boldsymbol{H}(w)\mathrm{E}[\boldsymbol{s}(q,w)\boldsymbol{s}(q,w)^{H}]\boldsymbol{H}(w)^{T}$$
$$=\boldsymbol{H}(w)\boldsymbol{H}(w)^{T} \qquad (19)$$

By imposing Tikhonov regularization techniques [47] to avoid the ill-posed problem, where it is well-known that regularization is an effective way to avoid the ill-conditioned matrix, the equation (19) becomes as follows:

$$\mathbf{R}(q,w)+c\mathbf{I}_N=\boldsymbol{H}(w)\boldsymbol{H}(w)^{T} \qquad (20)$$

Where $\mathbf{I}$ is an N x N identity matrix, and $c=m.\left(\mathrm{tr}\left(\mathbf{R}(q,w)\right)\right)$, it is regularization parameter with m is a positive constant and $\mathrm{tr}(.)$ is a trace operator of the estimation covariance matrix $\mathbf{R}(q,w)$. Note that the regularization method here just adds energy constraint in order to boost the covariance matrix to be a well-conditioned matrix. Therefore, the $\mathbf{R}(q,w)+c\mathbf{I}_N$ can be decomposed as

$$[\mathbf{R}(q,w)+c\mathbf{I}_N]=\mathbf{V}(w)\boldsymbol{\Lambda}(w)\mathbf{V}^H(w) \qquad (21)$$

where $\mathbf{V}(w)$ is a NxN matrix satisfying

$$\mathbf{V}(w)\mathbf{V}^H(w)=\mathbf{V}^H(w)\mathbf{V}(w)=\mathbf{I}_N \qquad (22)$$

and $\boldsymbol{\Lambda}(w)$ is an N x N diagonal matrix. So, from (22), the N x N matrix $\mathbf{H}(w)$ will be

$$\mathbf{H}(w)=\mathbf{V}(w)\boldsymbol{\Lambda}(w)^{-\frac{1}{2}}\mathbf{U}^H(w) \qquad (23)$$

where $\mathbf{U}(w)$ is a N x N full rank unitary matrix and $\mathbf{U}\mathbf{U}^H=\mathbf{I}_N$. However, the whitening step obtained matrix $\mathbf{V}(w)$ so that the $\mathrm{N \times 1}$ whitened data vector $\mathbf{Z}(\mathrm{q,w})$ has covariance of identity matrix $\mathbf{R}_{\mathrm{ZZ}}(\mathrm{q,w}) = \mathbf{I}_{\mathrm{N}}$, which can be obtained as follows:

$$\mathbf{Z}(\mathrm{q,w}) = \mathbf{\Lambda}^{-\frac{1}{2}}\mathbf{V}^{\mathrm{H}}\mathbf{x}(\mathrm{q,w}) \quad (24)$$

$$\mathbf{Z}(\mathrm{q,w}) = \mathbf{U}^{\mathrm{H}}(\mathrm{w})\mathbf{s}(\mathrm{q,w}) \quad (25)$$

The estimated source signals can be recovered with a linear Zero-Forcing (ZF) equalizer. Then the estimated KxT source vector is

$$\mathbf{y}(\mathrm{q,w}) = \mathbf{U}(\mathrm{w})\mathbf{Z}(\mathrm{q,w}) \quad (26)$$

After the preprocessing step, the estimation of the source signals $\mathbf{y}(\mathrm{q,w})$ reduces to determining the $\mathrm{N \times N}$ unitary matrix $\mathbf{U}(\mathrm{w})$ (rotation matrix).

*B. Step 2: Determining the rotation matrix (unitary matrix) $\boldsymbol{U}(w)$.*

One way of finding the rotational matrix $\mathbf{U}(\mathrm{w})$ is by maximizing the normalized fourth-order marginal cumulant (Kurtosis contrast) of the whitened data $\mathbf{Z}$ in (25). To estimate $\mathbf{U}(\mathrm{w})$ in (26), this paper exploits the statistical independence of the equalized source vector. More precisely, the unitary matrix $\mathbf{U}(\mathrm{w})$ will be estimated by utilizing the independent property of the estimated source vector at each frequency bin ($\mathbf{y}(\mathrm{q,w})$ ) in the normalized fourth-order marginal cumulant of whitened data $\mathbf{Z}(\mathrm{q,w})$ as follows:

$$\mathbf{K}(\mathbf{q},\mathbf{w}) = \frac{\mathbf{E}[|\mathbf{y}(\mathbf{q},\mathbf{w})|^4] - 2\mathbf{E}^2[|\mathbf{y}(\mathbf{q},\mathbf{w})|^2] - |\mathbf{E}[\mathbf{y}(\mathbf{q},\mathbf{w})^2]|^2}{\mathbf{E}^2[|\mathbf{y}(\mathbf{q},\mathbf{w})|^2]} \quad (27)$$

where $\mathrm{E}[\cdot]$ represents the expectation operator. Based on the deflation approach to ICA [30], one can estimate the *nth* source signal as follows

$$\mathbf{y}_n(\mathrm{q,w}) = \mathbf{u}_{\mathrm{n}}^{\mathrm{H}}(\mathrm{w})\,\mathbf{x}(\mathrm{q,w}) \quad (28)$$

where $(\cdot)^H$ represents the conjugate-transpose operator, $\mathbf{u}_n(w)$ is the *nth* column vector of the demixing matrix $\boldsymbol{U}(w)$ and $\boldsymbol{y}_n(q,w)$ is *nth* source signal at each $wth$ frequency bin and *qth* frame time. According to [1], [2], the column vector $\boldsymbol{u}_n(w)$ of the demixing matrix $\boldsymbol{U}(w)$ can be estimated for all users due to the batch adaptation by a gradient decent method as follows

$$\boldsymbol{u}_n^{l+1} = \boldsymbol{u}_n^{l} - \mu\Delta\boldsymbol{G}_n^{l} \quad (29)$$

where l denotes the iteration index, $\mathbf{u}_{\mathrm{n}}^{\mathrm{l}}$ is the *nth* column vector of the demixing matrix $\mathbf{U}(\mathrm{w})$ at $lth$ iteration and $\Delta\mathbf{G}_{\mathrm{n}}^{\mathrm{l}}$ is the gradient of the contrast measure that updates the demixing vector $\mathbf{u}_{\mathrm{n}}^{\mathrm{l}}$ in the demixing matrix $\mathbf{U}(\mathrm{w})$. Gradient function depends on the cost function that ICA would maximize /minimize in order to extract the source signal [5]. Herein, this paper refers to the use of the ICA techniques based on the kurtosis criterion, which is given in (27), as follows:

$$K(q,w) = \frac{E[|\mathbf{y}(q,w)|^4] - 2E^2[|\mathbf{y}(q,w)|^2] - |E[\mathbf{y}(q,w)^2]|^2}{E^2[|\mathbf{y}(q,w)|^2]} \quad (30)$$

Having the RobustICA's search-method of the kurtosis criterion in (30) in order to choose the optimal step size [6] as follows:

$$\mu_{\mathrm{opt}} = \arg \max_{\mu} |\mathrm{K}(\boldsymbol{y}(q,w) + \mu\mathbf{g}(q,w))| \quad (31)$$

where $\mathbf{g}$ is the gradient of Kurtosis contrast $\mathrm{K}(.)$. One can easily choose the optimal step size $\mu_{\mathrm{opt}}$ based on one of the algebraic methods instead of using the exact line search as in [13], [14] to avoid the intensive computation and other limitations as in [6]. Therefore, it is easy to find the global optimum step size $\mu_{\mathrm{opt}}$ for the criteria that can be expressed as a polynomial function of $\mu$ due to its roots, e.g. the criteria kurtosis [6], the constant modulus [13] and the constant power [2]. Therefore, the RobustICA performs an optimal step-size of estimating the *nth* source signal, based on optimization, for *lth* iteration, $wth$ frequency bin, and $qth$ frame as follows:

- *Step1) initialize value for the weight vector* $\boldsymbol{u_n}(w)$
- *Step2) Compute the optimal step size polynomial coefficients. For Kurtosis contrast, the optimal step size polynomial is given by*

$$p(\boldsymbol{u_n}(w)) = \sum_{n=0}^{4} a_n\mu^n \quad (32)$$

*where the coefficients $a_n$ can be obtained at each iteration by the observed signal block and the current values of $w$ and $g$. Details can be found in [6].*

- *Step 3) Extract the optimal step size polynomial root $\mu^n$. The root can be obtained by using the Ferrari's formula as in [48].*
- *Step 4) Select the optimal step size polynomial root $\mu^n$ as follows*

$$\mu_{opt} = arg \max_{\mu} |K(\boldsymbol{y}_n^l(q,w) + \mu\boldsymbol{g}_n^l(q,w))| \quad (33)$$

- *Step 5) Find the updated weighed vector*

$$\boldsymbol{u}_n^{l+1} = \boldsymbol{u}_n^{l} - \mu_{opt}\boldsymbol{g}_n^{l} \quad (34)$$

*where $g_n^l$ is the nth gradient of Kurtosis contrast $K(.)$ at lth iteration.*

- *Step 6) Normalize and update the weight vector*

$$\boldsymbol{u}_n^{l+1} = \frac{\boldsymbol{u}_n^{l+1}}{\|\boldsymbol{u}_n^{l+1}\|} \quad (35)$$

*where $\|\boldsymbol{u_n}\|$ is a norm of $\boldsymbol{u_n}$.*

- *Step 7) Go back to step 2 until the convergence.*

To prevent locking onto a previously extracted source, or when the old and new vectors $\boldsymbol{u_n}, \boldsymbol{u_{n+1}}$ are in the same direction, the learning converges and their absolute dot-product value reaches close to 1. Thus, owning the deflation method proposed in [30] avoids different vectors from converging at the same maxima. However, each vector of $\mathbf{U} = \{\mathbf{u}_1, \mathbf{u}_2, \ldots, \mathbf{u}_{\mathrm{n}}\}$ needs to be orthogonalized before each iteration. Based on the Gram-Schmidt orthogonalization, the deflation scheme estimates each independent component at each iteration step. Gram-Schmidt orthogonalization of $(n+1)th$ component can be expressed as follows

$$\mathbf{u}_{\mathrm{n+1}}^{\mathrm{l+1}} = \mathbf{u}_{\mathrm{n+1}}^{\mathrm{l}} - \sum_{\mathrm{n=1}}^{\mathrm{N}} \left(\mathbf{u}_{\mathrm{n+1}}^{\mathrm{l}}{}^{\mathrm{H}}\mathbf{u}_{\mathrm{n}}^{\mathrm{l}}\right)\mathbf{u}_{\mathrm{n}}^{\mathrm{l}} \quad (36)$$

$$\mathbf{u}_{\mathrm{n+1}}^{\mathrm{l+1}} = \frac{\mathbf{u}_{\mathrm{n+1}}^{\mathrm{l+1}}}{\|\mathbf{u}_{\mathrm{n+1}}^{\mathrm{l+1}}\|} \quad (37)$$

where a new weight vector $\mathbf{u}_{\mathrm{n+1}}$ is obtained by subtracting the vector projected from the old weight vector.

The following steps summarize the presented algorithm procedure:

- *Start*
- *Perform the time-frequency representation as in (4).*
- *For each frequency bin $w = 1, \ldots, \frac{T}{2}$*
- *Pre-processing of the observed data $\boldsymbol{x}(q,w)$ and imposing the Tikhonov regularization parameter to avoid the ill-conditioning problem of the covariance matrix and to mitigate the performance degradation.*

$$\alpha = m.tr(E[\boldsymbol{x}(q,w)\boldsymbol{x}^H(q,w)]) \quad (38)$$

*where $m$ is a positive constant and $tr(\cdot)$ represents the trace of the estimation covariance matrix of the observation signals.*

- *Initialize $N$ x $N$ matrix $\boldsymbol{W}$ (equals identify matrix $I$), where $N$ is the number of sources.*
- ***For** each user $n = 1, \dots, N$*
- *Initialize $\boldsymbol{u}_n$ column vector of the demixing matrix $\boldsymbol{U}$*
- ***While loop***
- *Evaluate $\mathbf{y}_n(q,w)$ in (13)*
- *Select the optimal step size polynomial root $\mu_{opt}$ in (33)*
- *Update weighed vector in (34)*
- *Do the orthogonalization and normalization in (36) and (37), respectively*
- *Find nth users:*

$$\boldsymbol{y}_n(q,w) = \boldsymbol{u}_n^H(w)\, \boldsymbol{z}(q,w) \quad (39)$$

- *Do deflation by subtracting the estimated nth source contribution to the $\boldsymbol{z}(q,w)$ as follows [30]:*

$$\boldsymbol{z}_{n+1}(q,w) = \boldsymbol{z}_n(q,w) - \boldsymbol{h} * \boldsymbol{y}_n(q,w) \quad (40)$$

*Where $\boldsymbol{h}$ is the symbol direction estimated via least squares, and it is given by*

$$\boldsymbol{h} = \frac{\mathbf{z}_k(q,w).\mathbf{y}_k^H(q,w)}{\mathbf{y}_k(q,w).\mathbf{y}_k^H(q,w)} \quad (41)$$

- *Check the convergence point. if so, End **while loop**, otherwise, go back until the convergence.*
- *Save $\boldsymbol{u}_n$ in the $\boldsymbol{U}(w)$.*
- *End **for loop**.*
- *Save the demixing matrix $\boldsymbol{U}(w)$*
- *End w loop.*

## IV. Scaling and Permutation Ambiguities

Assume $\mathbf{U}(\mathrm{w})$ is the rotational matrix that is computed at each bin. The least square estimation of the mixing matrix $\mathbf{H}(\mathrm{w})$ is given by

$$\mathbf{H}_{\mathrm{LS}}(\mathrm{w}) = \mathbf{x}(\mathrm{q},\mathrm{w})\mathbf{y}'(\mathrm{q},\mathrm{w})(\mathbf{y}(\mathrm{q},\mathrm{w})\mathbf{y}'(\mathrm{q},\mathrm{w}))^{+} \quad (42)$$

where

$$\mathbf{y}(\mathrm{q},\mathrm{w}) = \mathbf{U}(\mathrm{w})\mathbf{x}(\mathrm{q},\mathrm{w}) \quad (43)$$

However, one can express the estimated mixing matrix $\mathbf{H}_{\mathrm{LS}}(\mathrm{w})$ in terms of the perfect mixing matrix $\mathbf{H}(\mathrm{w})$ as follows:

$$\mathbf{H}_{LS}(w) = \mathbf{H}(w)\mathbf{D}^{-1}(w)\mathbf{\Gamma}^{-1}(w) \quad (44)$$

where $\mathbf{D}(w)$ is an unknown diagonal matrix and $\mathbf{\Gamma}(w)$ is an unknown permutation matrix. Therefore, we have to estimate $\mathbf{D}(w)$ and $\mathbf{\Gamma}(w)$ matrices to solve the scaling and permutation ambiguities.

### A. Estimation of the diagonal matrix $\boldsymbol{D}(w)$

Several methods to compensate the scale ambiguity have been proposed in the literature. Thus, we choose to estimate the diagonal matrix $\mathbf{D}(\mathrm{w})$ using the minimal distortion principle [3], [22]. The $\mathbf{D}(\mathrm{w})$ is given in [3] as

$$\mathbf{D}(w) = diag[\boldsymbol{A}\mathbf{H}_{LS}(w)] \quad (45)$$

$$\mathbf{D}(w) = diag[\mathbf{A}\boldsymbol{x}(q,w)\boldsymbol{y}'(q,w)(\boldsymbol{y}(q,w)\boldsymbol{y}'(q,w))^{+}] \quad (46)$$

where $\mathrm{A} \in \mathbb{R}^{\mathrm{N\,x\,M}}$ is a matrix in which all its entries are $1/_{\mathrm{M}}$, and where $M$ is the number of observations whereas $N$ is the number of sources, and $\mathrm{diag}[\mathbf{C}]$ returns a matrix that contains the diagonal elements of matrix $\mathbf{C}$ and sets the other non-diagonal elements of matrix $\mathbf{C}$ zeros.

The interpretation of (46), in a sense of perfect separation, is that each estimated source averages along the sensors in the sense of all other sources have turned off. In other words, the Minimal Distortion Principle assumes that the *nth* source is scaled with respect to the image at the nth microphone [40]. Therefore, the rescaled source signals can be expressed as follows:

$$\boldsymbol{y}^{rescaled}(q,w) \cong \mathbf{D}(w)\mathbf{x}(q,w) \quad (47)$$

$$\boldsymbol{y}^{rescaled}(q,w) \cong diag[\mathbf{A}\mathbf{x}(q,w)\mathbf{y}'(q,w)(\mathbf{y}(q,w)\mathbf{y}'(q,w))^{+}]\mathbf{x}(q,w) \quad (48)$$

### B. Estimation of the permutation matrix $\boldsymbol{\Gamma}(w)$

Despite the fact that this section estimates the permutation matrix $\mathbf{\Gamma}(\mathrm{w})$ proposed in several current works in the literature, estimating the permutation matrix is still considered a very challenging problem that needs to be addressed. Assume that we have N source signals which are presented in the BSS problem; then there are N factorial times the possible permutations at each bin, which yields a complex combinational problem.

There are several previously mentioned techniques used to solve the permutation problem in the literature [3], [22]. In this paper, we will review and evaluate them in terms of computational complexity and performance.

One can divide these methods into two main solution groups, which solve the permutation ambiguity in the frequency domain as follows:

- *Group based on geometric information, such as Time Direction of Arrivals (TDOA) and Direction of Arrivals (DOA) [3], [21], [22], [40].*
- *Group based on clustering-based techniques [22], [31], [32], [35].*

Many of these techniques are based on geometric information, such as estimation of the direction of arrival (DOA) and Time difference of Arrival (TDOA) as in [22], [27]. Other techniques depend on the coherence of the un-mixing filter coefficients. In other words, these techniques take advantage of some prior knowledge about mixing filters and restrict the mixing matrix H(w) to be continuous in frequency domain. Furthermore, in [34], Parra imposes smoothness to the de-mixing filter values in the frequency domain. Also, a restriction is made with the frequency domain update rule so that it is associated with the limited length filter in the time domain. Such a restriction may not be considered sophisticated, especially in a case of reverberant environment, since it is necessary to have a long length filter to cover all reverberations. Although it can be avoided by choosing a large frame size, it still causes more overall complexity, especially, when the short mixtures are available. In terms of the properties of speech, there are other categories, which have been proposed in literature, to estimate the permutation matrix and make the spectral alignment.

The most common is based on the inter-frequency correlation of speech envelopes [33], [36]. The inter-frequency correlation technique exploits the nature of speech production, where it's known that all spectral components of speech signals increase as the talker speaks louder. In that sense, several weighted techniques and criteria have been proposed to impose the frequency-coupling between the adjacent frequency bins. For more details, see [3], [22]. Although these techniques perform well in the simulations, they are not sophisticated when they are applied to a real recording room. They suffer from propagation error or delays. For example, if an error occurs at a certain frequency bin, it may increase the possibility that it will occur again at the following frequency bins. Therefore, in the literature [3], [22], [32], researchers avoid propagation error by estimating a frequency-independent reference profile, which is called a centroid, due to using a clustering-based method for each separated source. They then structure the $N$ frequency-dependent profiles such that they are all matched with a different frequency-independent reference profile at each frequency bin.

The main steps of the clustering-based techniques are as follows:

- *Define the quantities that are used in the clustering, such as the signal envelopes of the source profiles, the log-power of the source profiles, etc.*
- *Choose the measure that is used to determine the matching level between the centroids and the profiles, such as correlation, distance, etc.*
- *Choose the cluster technique.*

In [21], [28], the profile $\boldsymbol{\Psi}_{\mathrm{n}}(\mathrm{q},\mathrm{w})$ of a separated signal $\mathbf{y}_{\mathrm{n}}$ is chosen as the envelope of the separated source $\mathbf{s}_{\mathrm{n}}$ where $\boldsymbol{\Psi}_{\mathrm{n}}(\mathrm{q},\mathrm{w}) = |\mathbf{y}_{\mathrm{n}}(\mathrm{q},\mathrm{w})|$. In [22], Authors are chosen for the profile $\boldsymbol{\Psi}_{\mathrm{n}}(\mathrm{q},\mathrm{w})$ of a separated signal $\mathbf{y}_{\mathrm{n}}$ to be a certain dominance measure. Whereas, in [38], the profile $\boldsymbol{\Psi}_{\mathrm{n}}(\mathrm{q},\mathrm{w})$ of a separated signal $\mathbf{y}_{\mathrm{n}}$ is defined to be its centered log-power spectral density where the log-power profile is given as follows:

$$\boldsymbol{\Psi}_{\mathrm{n}}(\mathrm{q},\mathrm{w}) = \log\left[\mathbf{U}_{\mathrm{n},:}(\mathrm{w})\mathbf{R}_{\mathrm{xx}}(\mathrm{q},\mathrm{w})U_{\mathrm{n},:}^{\mathrm{H}}(\mathrm{w})\right] \quad (49)$$

In clustering-based approaches, the length of the profiles $\mathrm{T_f}$ is also an important parameter in terms of accuracy, especially for short signals. This work is essentially going to set up the profiles for the overlapping frames over the whole signal. Once we construct the profiles of the separated signals, then we compute the centroids in order to perform the clustering. The clustering-based technique is essentially based on the assumption that profiles coming from the same source at different frequency bins still have more match level than those coming from other sources. Actually, the most common methods to associate each source profile to a centroid at each frequency bin are based on 1) maximizing correlation measures [69], [70] and 2) minimizing distance measures across the $N$ factorial times of the possible permutations at each frequency bin [38]. However, authors employ the iterative techniques to update the centroids and the permutation matrices. In other words, they update the centroids first, and then they permute the source profiles to each desired centroid and match them together using one of the two previous measures, i.e. distance [38] or correlation in [21], [28] and [31].

In spite of the fact that the aforementioned iterative methods perform well, they tend to be significantly more expensive in terms of cost and computational complexity since they have the N factorial times of the possible permutations at each frequency bin. To avoid this drawback in the aforementioned iterative methods in [32], Nion et al. propose a more efficient modification of the clustering strategy, which is updating the whole permutation matrices and centroids simultaneously. In other words, the update of the centroids and permutation matrices are not interleaved. Thus, their modification has improved these iterative methods in terms of computational complexity.

Their methods can be summarized as follows:

*Step 1. Determine the centroids and compute them.*

Consider the $\mathrm{N} \times \mathrm{T_f}$ matrix $\boldsymbol{\mathcal{F}}(\mathrm{w})$ that is structured from the N profiles $\boldsymbol{\Psi}_{\mathrm{n}}(\mathrm{w}),\ \forall\ \mathrm{n} = 1,\dots,\mathrm{N}$. One can extend the $\mathrm{N} \times \mathrm{T_f}$ matrix $\boldsymbol{\mathcal{F}}(\mathrm{w})$ to the $\mathrm{FN} \times \mathrm{T_f}$ matrix $\mathbf{G}(\mathrm{w})$ by concatenating the matrices $\boldsymbol{\mathcal{F}}(\mathrm{w})\ \forall\ \mathrm{w} = 1,\dots,\mathrm{F}$. In order to enforce the FN profile points in matrix $\mathbf{G}(\mathrm{w})$ varying smoothly with time, we have to encounter the computation of the profiles for overlapping frames. Hereafter, we just need to classify these FN profile points into N clusters due to applying the k-mean algorithm on the $\mathrm{FN} \times \mathrm{T_f}$ matrix $\mathbf{G}(\mathrm{w})$ to carry out a frequency-independent $\mathrm{N} \times \mathrm{T_f}$ centroid matrix $\mathbf{M} = \left[\mathbf{m}_1^{\mathrm{T}}, \mathbf{m}_2^{\mathrm{T}}, \dots, \mathbf{m}_{\mathrm{N}}^{\mathrm{T}}\right]^{\mathrm{T}}$. The centroid matrix is structured by summing all the points within a cluster, which have attained a minimum distance regarding the centroid cluster.

*Step 2. Estimate the permutation matrices.*

In the previous step, we reduced the computational processes to find that the $\mathrm{N} \times \mathrm{N}$ permutation matrix $\boldsymbol{\Gamma}(\mathrm{w})$, subject to $\mathbf{G}(\mathrm{w})\boldsymbol{\Gamma}(\mathrm{w})$, matches the frequency-independent $\mathrm{N} \times \mathrm{T_f}$ centroid matrix $\mathbf{M}$ at each frequency bin. Therefore, one can choose to minimize the distance that is given in [38] as follows

$$\min_{\boldsymbol{\Gamma}(w)} \|\boldsymbol{M} - \boldsymbol{G}(w)\boldsymbol{\Gamma}(w)\|_F^2 \ \ \forall\, w = 1,2,\dots F \quad (50)$$

Or one can chose the correlation criteria that is given by [21], [22] [31] as follows

$$\max_{\boldsymbol{\Gamma}(w)} \sum_{n=1}^{N} \Phi\langle \boldsymbol{m}_n, [\boldsymbol{G}(w)\boldsymbol{\Gamma}(w)]_{:,n} \rangle \ \ \forall\, w = 1,2,\dots F \quad (51)$$

where $\Phi\langle \cdot, \cdot \rangle$ is the correlation coefficient.

In terms of performance, the first group generally does better than the second group, especially at the small data sample available. But it is not optimal in a practical sense, since we don't usually have geometric information about the real environmental conditions. In that sense, the second group performed better than first group, especially if we have a large sample set of data, because they are based on the clustering-based techniques (i.e.: correlation, distance, etc.) and they are more robust to real-world scenarios. For more details, refer to [3] [22].

## V. Experiments' Results

In this section, Monte Carlo Simulations are carried out. It is assumed that the number of sources is equal to the number of observation “sensors”. The experiments have been carried out using the MATLAB software on an Intel Core i5 CPU 2.4-GHz processor and 4G MB RAM. We examine the performance of the RobustICA-based algorithm developed in this paper. The time-frequency representation of the observed data is computed as explained in section II due to the Short-Time-Fourier-Transform. Then, for each frequency bin, we find the demixing matrix.We will solve the scale and permutation ambiguities based on the aforementioned techniques.

To help explain this process, we divided this section into two subsections. First, we illustrate the performance of the RobustICA-based algorithm with different permutation methods in the literature [3], [22]. We study the effect of the type of the windows on the performance of the presented algorithm as well as the effect of overlapping parameter. Second, we provide the performance of the presented algorithm in two real-world scenarios that are generated in adverse conditions by F. Nesta, in [40], and compare it with other state-of-the-arts in [40], [20], and [34] and [38], labeled as “RR-ICA”, “IVA”, “Parra”, and “Pham”, respectively. In this paper, we evaluate the performance of the presented algorithm due to the BSS_EVAL toolbox, which is proposed in [49]. We use time-invariant filters of 1024 taps to represent the signal-to-interference ratio (SIR) and source-to-distortion ratio (SDR).

## *A. Section 1*

In this subsection, we study the computational complexity and the performance of the presented algorithm based on several criteria to solve the scale and permutation ambiguities in the frequency domain BSS problem.

Let’s define these criteria as follows:

- *Method 1 is the RobustICA-based algorithm with clustering of envelope profiles with a distance measure iterative procedure [22], [31].*
- *Method 2 is the RobustICA-based algorithm with clustering of log-power profiles with a correlation measure iterative procedure [22], [50].*
- *Method 3 is the RobustICA-based algorithm with clustering of envelope profiles with a distance measure kmeans procedure [35], [32].*
- *Method 4 is the RobustICA-based algorithm with clustering of log-power profiles with a correlation measure kmeans procedure [22], [32].*
- *Method 5 is the RobustICA-based algorithm with clustering of dominance-profiles with a correlation measure iterative procedure [22].*
- *Method6 is the RobustICA-based algorithm with clustering of dominance-profiles with a correlation measure iterative kmeans procedure [22], [32].*

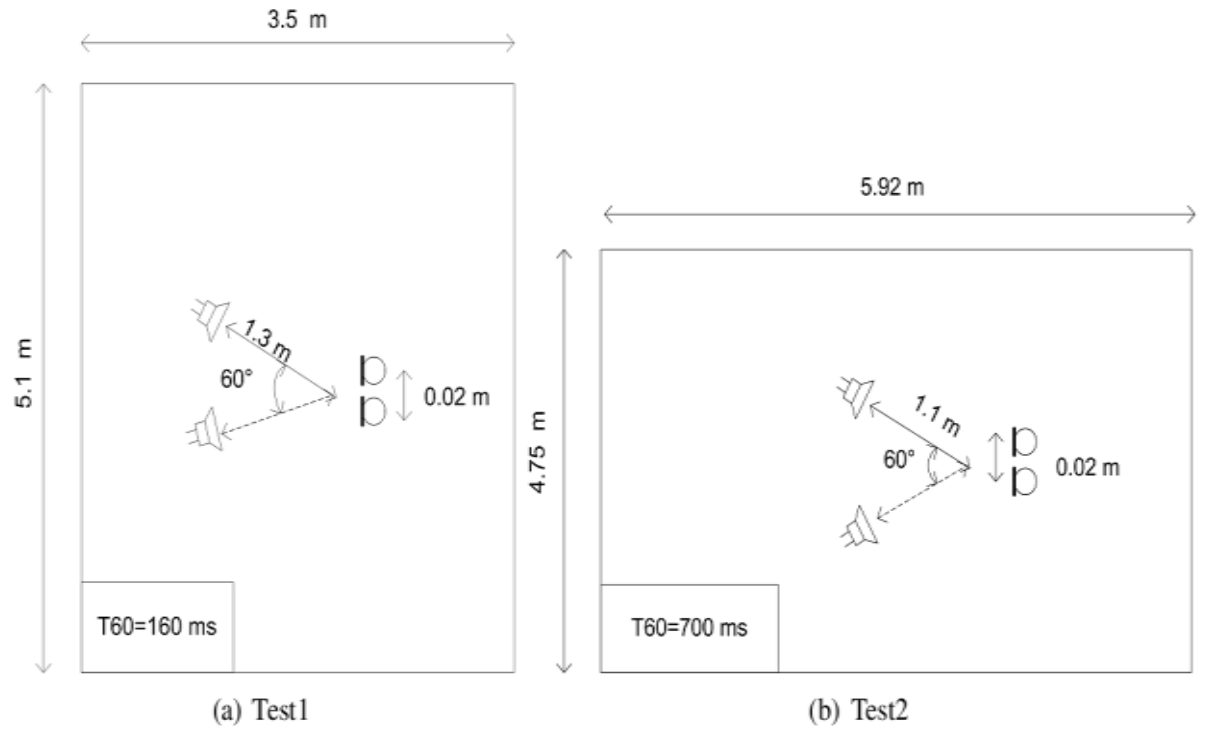

Fig. 1. Configuration of the two experimental setups that were conducted by Francesco Nesta1 in [40]:, a) room is characterized for Test1, b) class-room is characterized for Test2

In this section, we have used real world recordings, drawn from the experiments which were conducted in [40] named Test1. We would like to thank the authors who provided these recordings on their website “http://bssnesta.webatu.com/testhscma.html”. The two sources were recorded at $f_s = 16\,kHz$ with two microphones spaced $d = 0.02\,m$ apart to avoid spatial aliasing. The chosen room was characterized by a moderate reverberant time of $160\,ms$. The room had dimensions of $(3.5m\ x\ 5.1m\ x\ 2.6m)$ as shown in Fig. 1. The signal duration was fixed at 9 sec.

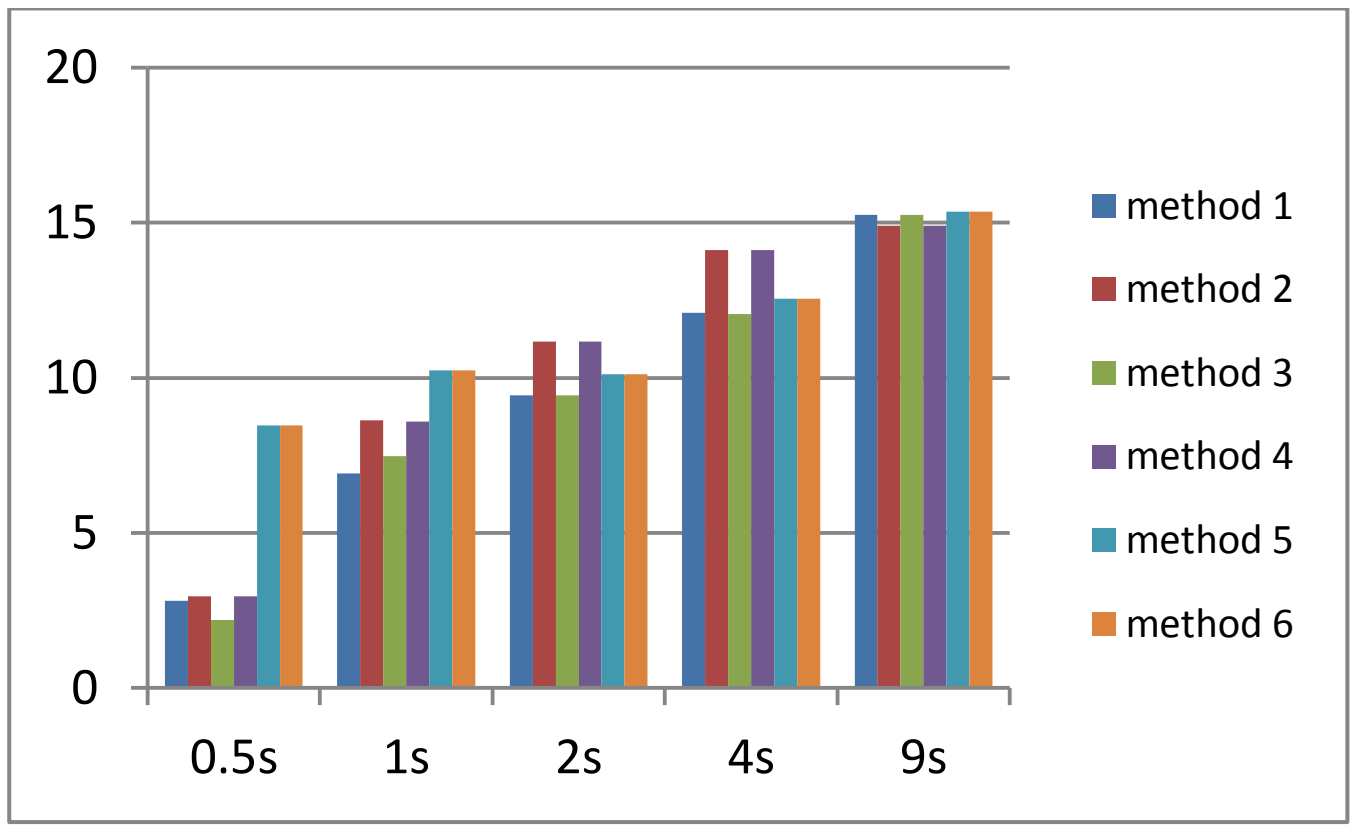

Fig. 2. Results obtained in Test1 experiments. The SIR performance of the presented algorithm with various permutation solvers

In fig. 2 and fig.3, we show the performance of the RobustICA-based algorithm with various aforementioned techniques of permutation solvers in terms of the SIR and SDR, respectively. In comparison, we notice that the dominance-profiles provide more robustness in terms of the signal’s length, although the envelope profiles are more sensitive to the signal’s length than the log-power profiles. Moreover, the dominance-profiles’ approach with the iterative procedure has the same performance as with the kmean procedure. Also, Fig.4 shows the corresponding CPU time of each permutation method that need to solve the permutation ambiguity.

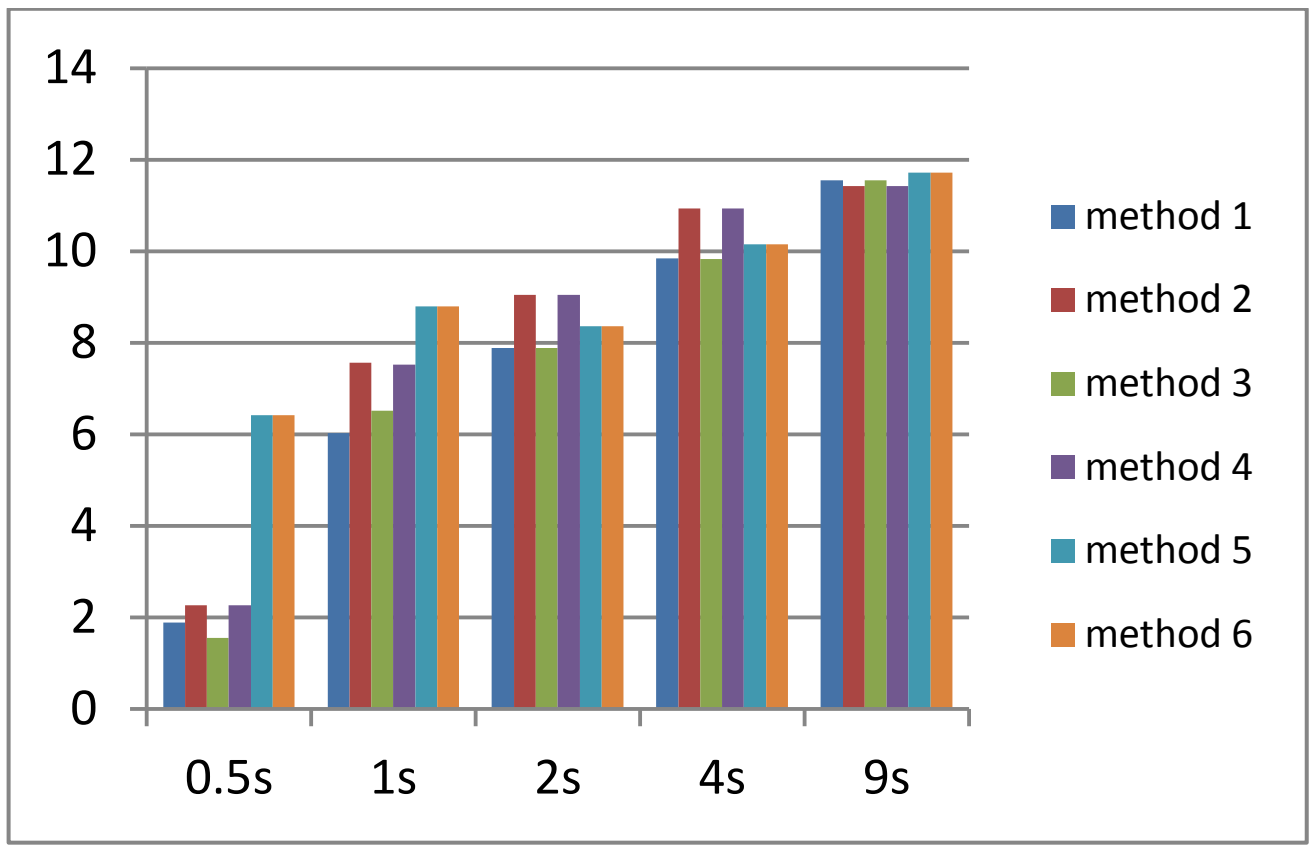

Fig. 3. Results obtained in Test1 experiments. The SDR performance of the presented algorithm with various permutation solvers

*Based on these observations, we use the dominance-profiles approach* with the iterative procedure after the RobustICA-based algorithm in the rest of these experiments. In Fig. 5, we illustrate the impact of the window's types on the performance of the proposed algorithm in terms of SIR and SDR respectively. And, we test the performance of the presented algorithm versus the overlapping parameter as shown in Fig 6.

The best performance of the presented algorithm was achieved during the certain range of the overlapping percentage. Therefore, based on these results, *we use the 0.65 overlapping parameters with Hamming window type*.

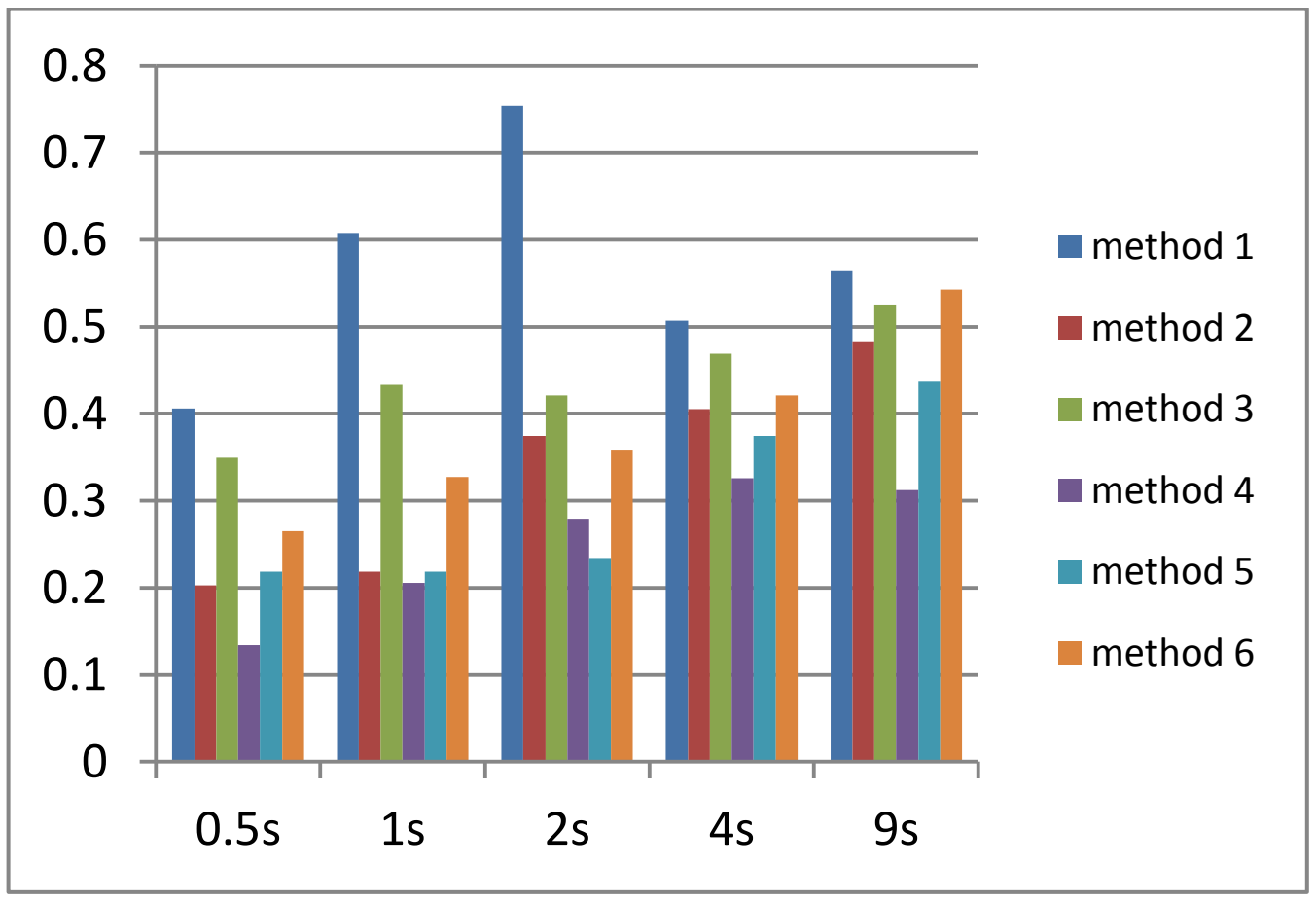

Fig. 4. Corresponding CPU time for each method.

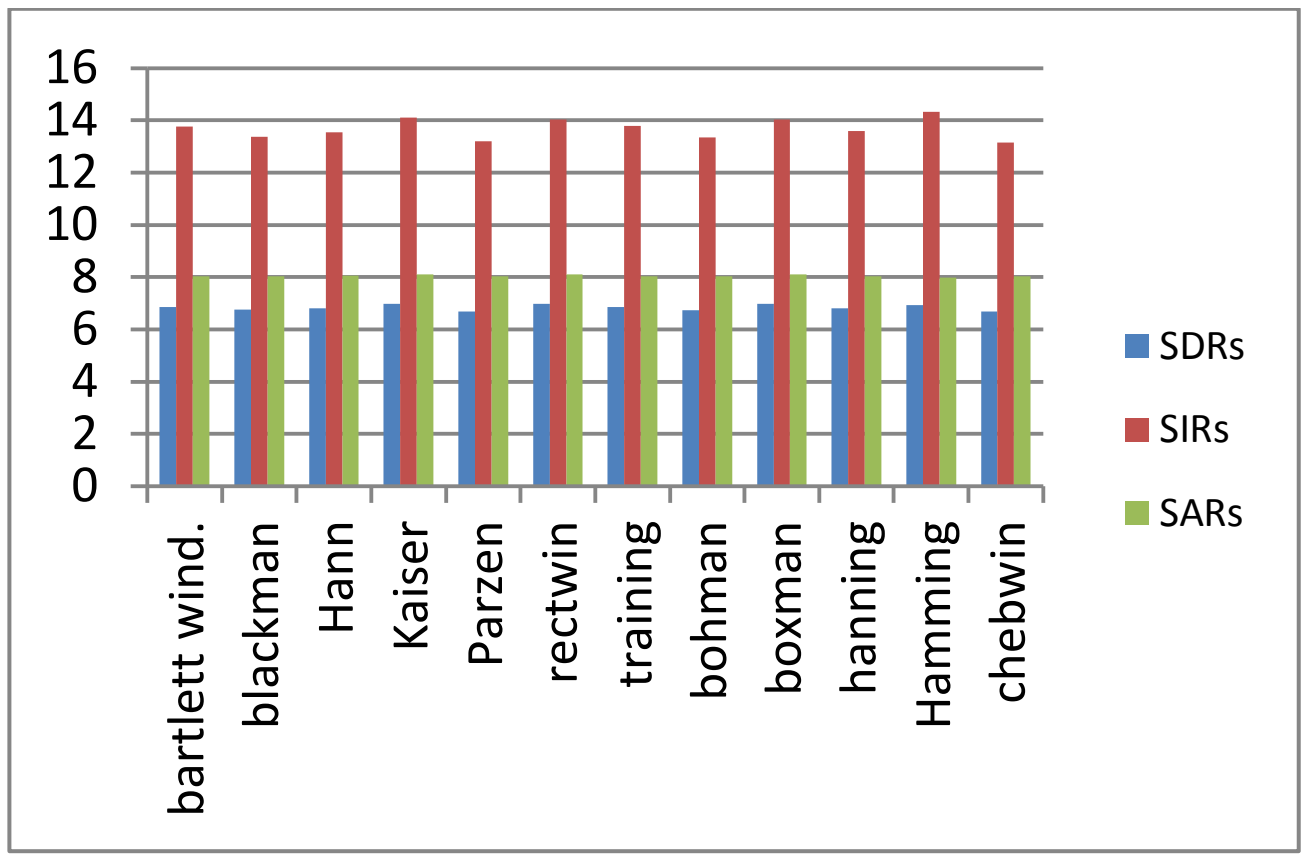

Fig. 5. Results obtained in Test2 experiments. The SIR performance of the presented algorithm with various window types

## B. Section 2

In this section, we perform the separation of the two mixture observations that consist of two sources. We have used the two tests "Test1 and Test2" of the real world recordings, drawn from the experiments that were conducted in [40] (see Fig. 1). Test2 uses the real world recordings of adverse reverberant conditions, as in Fig 1. The room is a reverberant class-room with dimensions *4.75 m Length* x *5.92 m Width* x *4.5 m Height*. The reverberation time is around 700 milliseconds $T_{60} = 700$ms. The signal duration was fixed to be 9 sec. After we got the demixing matrix **U** for each frequency bin, we used the Inverse Fourier Transform to obtain the mixing matrix in the time domain.

- *The independent vector analysis IVA [14] used with step size 0.1 and number of iterations is 1000.*
- *Parra's method [34], used with number of iterations is 1000.*
- *Pham's algorithm [35] and [38], used with FFT overlapping, equals* 75% *and a window size equal to 5.*
- *RR-ICA algorithm reported in [40].*

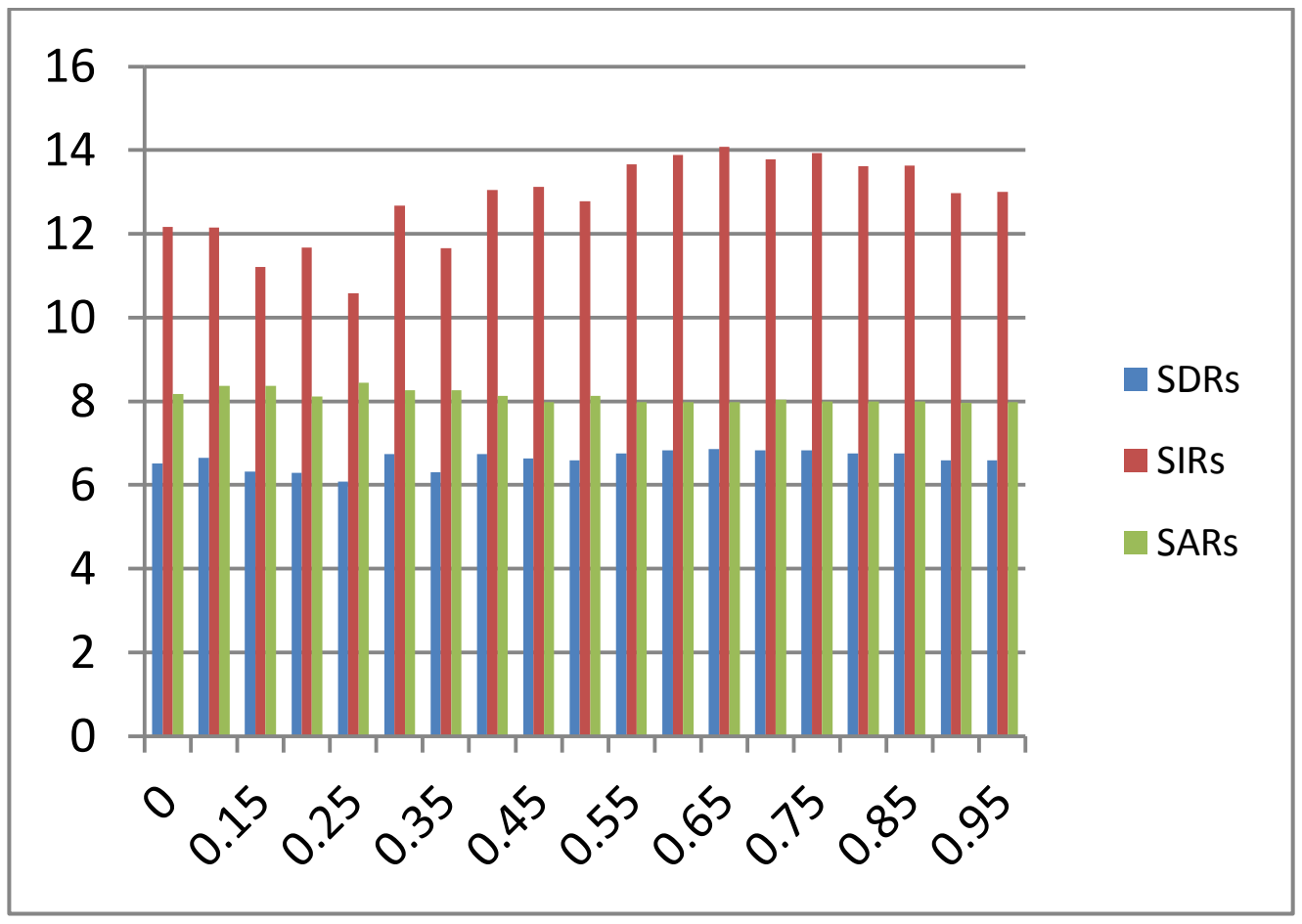

Fig. 6. Results obtained in Test2 experiments. The SIR performance of the presented algorithm with various overlap ratios

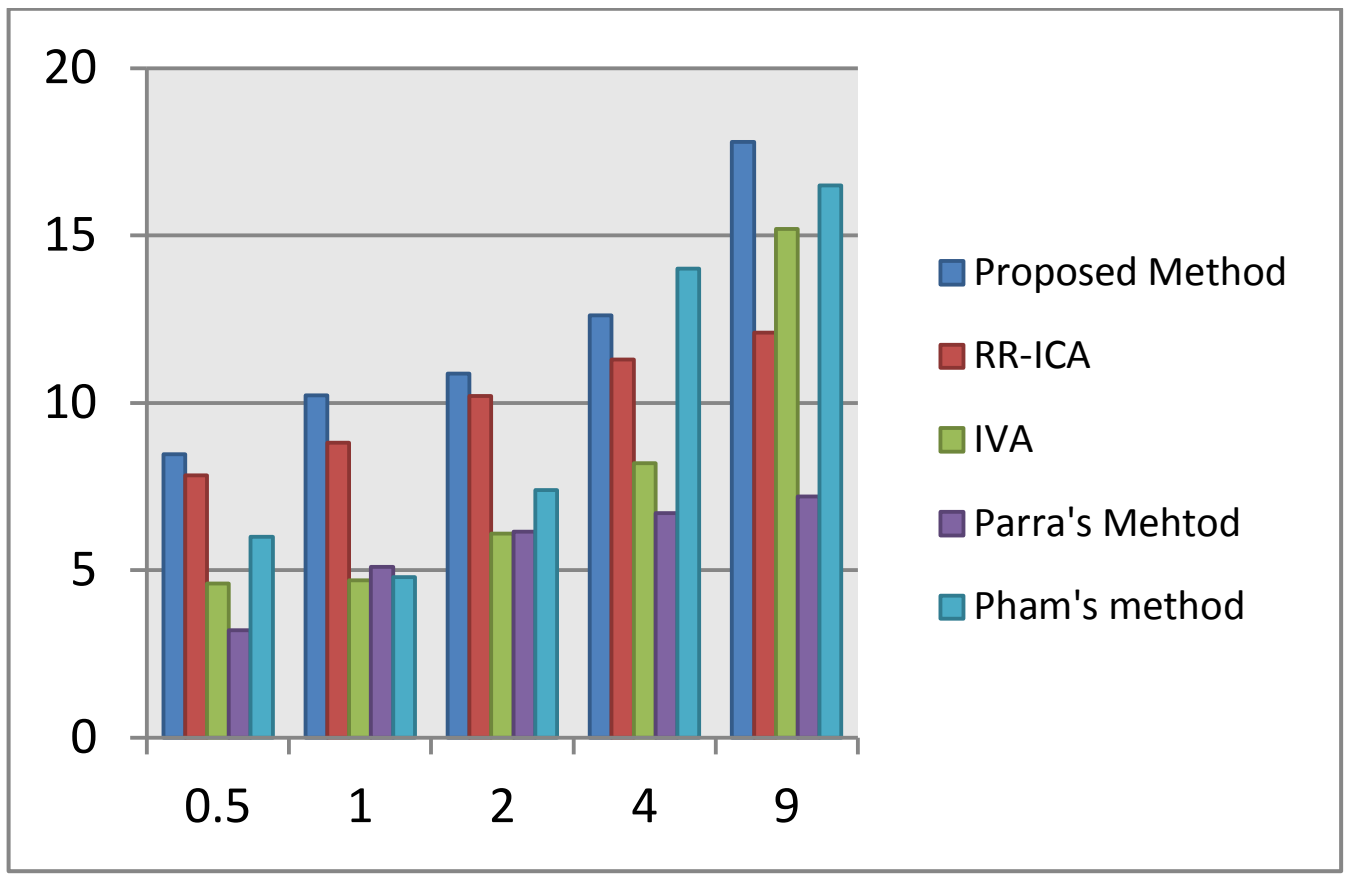

Fig. 7. Results obtained in the Test1 experiments. Best performance is reported in terms of SIR, by applying the given algorithms with different signal lengths

Fig 7 & 8 and 9 & 10, show the summary analysis of the presented algorithm versus other algorithms for the Test1 and Test2 configurations, respectively. These graphs report the best performance of each algorithm over the FFT size. Obviously, the RobustICA-based algorithm outperforms the other algorithms for any signal length in terms of SIR and SDR.

Moreover, in Fig. 11, we illustrate the impact of the FFT length on the performance of the proposed algorithm in terms of SIR. Clearly, the presented algorithm performs well, especially during reasonable FFT length in regard to other corresponding algorithms as shown in Fig.11.

Based on these results, one can show that the presented algorithm is stable in terms of the high reverberation environment and variations of the observations' parameters. Furthermore, the presented algorithm performs well in terms of stability and speed convergence. Owning the optimal step size, deflation and regularization techniques makes the presented algorithm more robust and allows it to perform well even in adverse conditions.

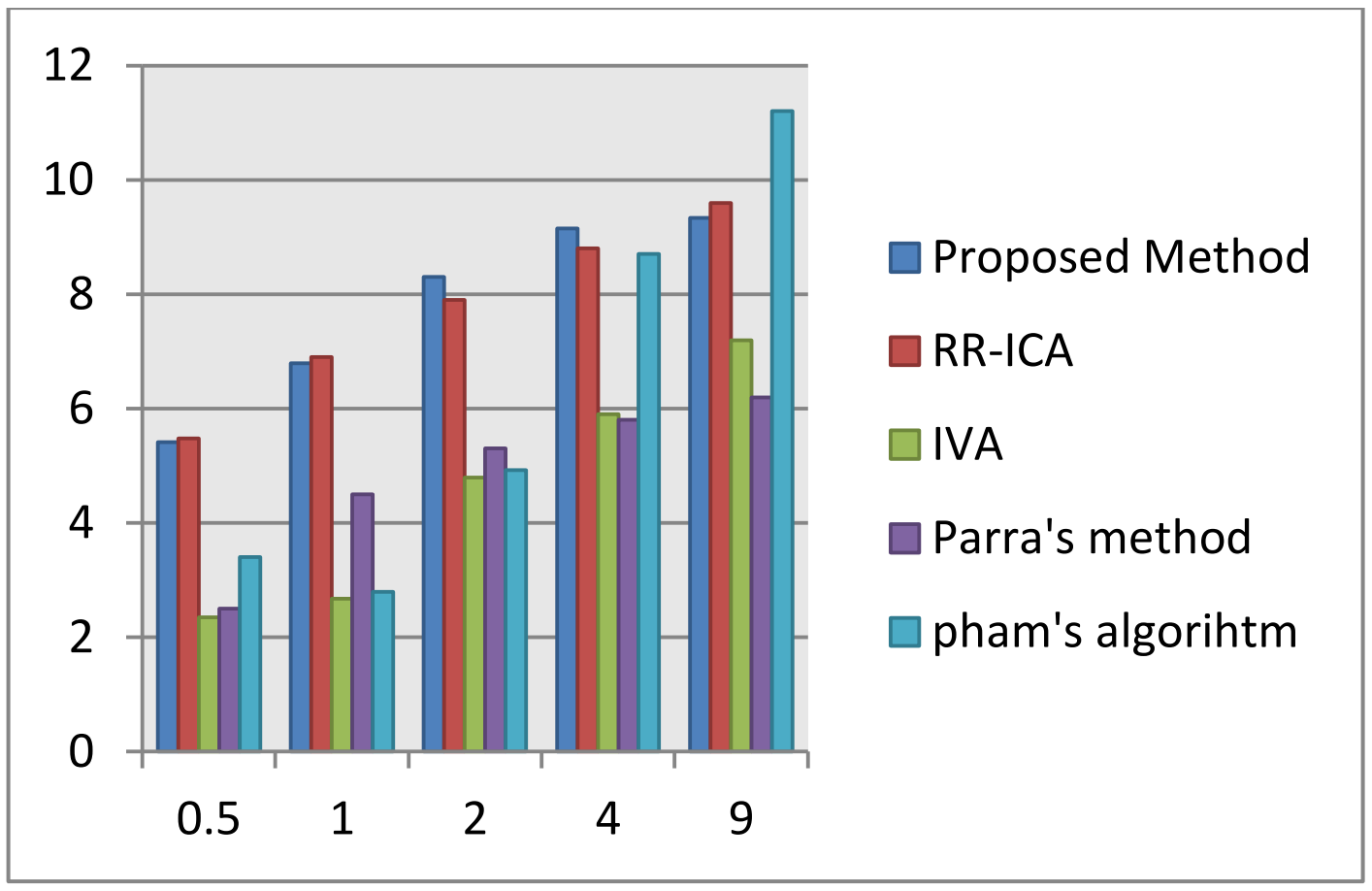

Fig. 8. Results obtained in the Test1 experiments. Best performance is reported in terms of SDR, by applying the given algorithms with different signal lengths

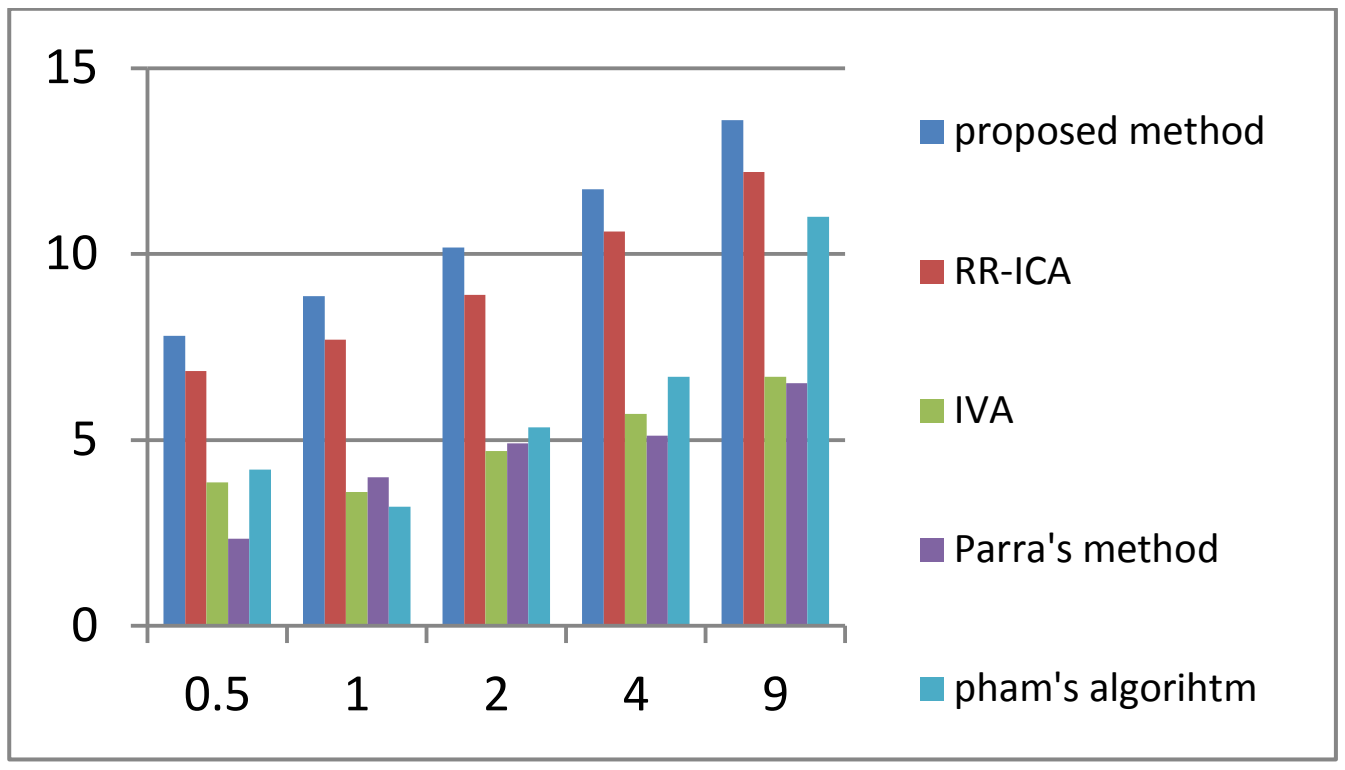

Fig. 9. Results obtained in the Test2 experiments. Best performance is reported in terms of SIR, by applying the given algorithms with different signal lengths

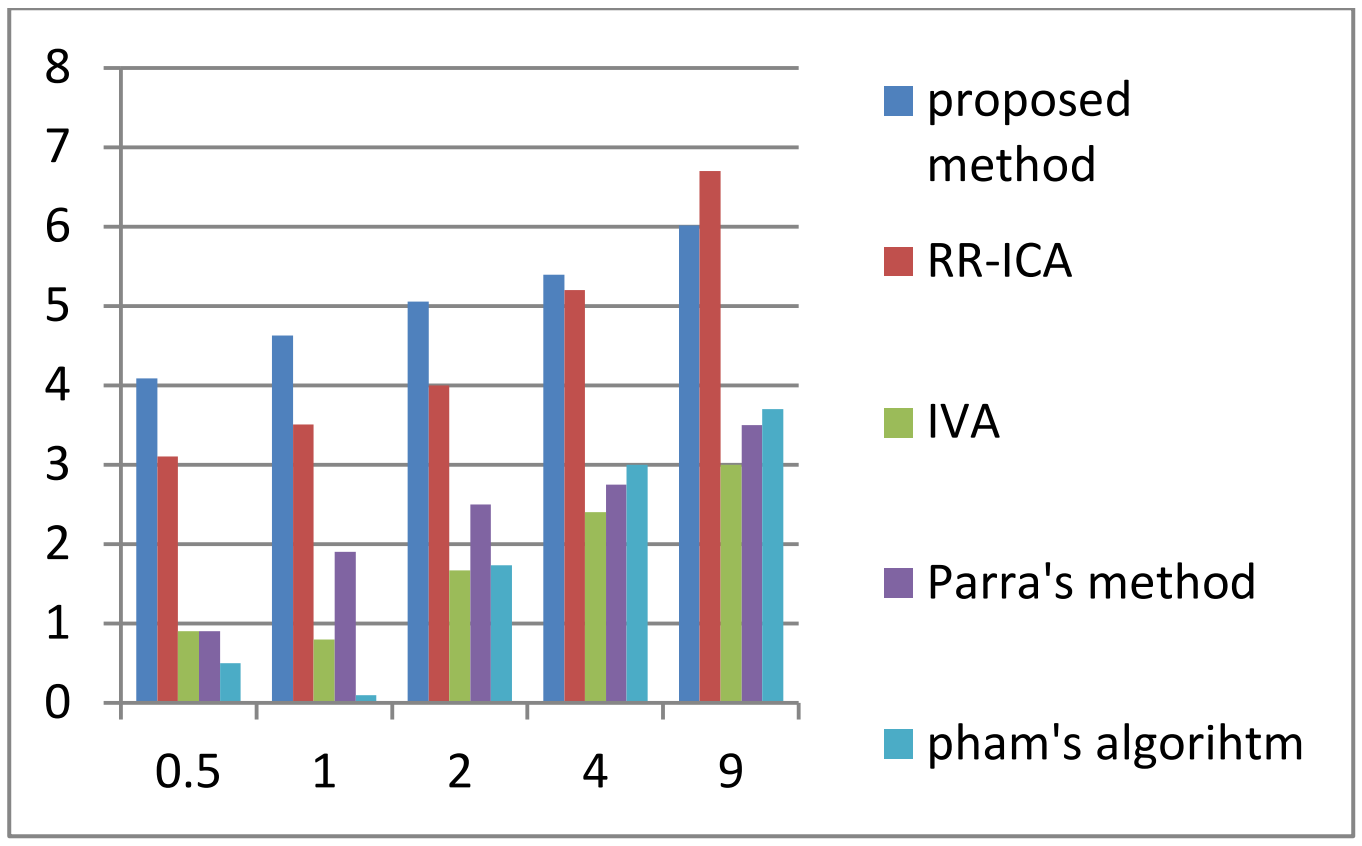

Fig. 10. Results obtained in the Test2 experiments. Best performance is reported in terms of SDR, by applying the given algorithms with different signal lengths

## VI. CONCLUSION

This paper presented the RobustICA-based algorithm to solve the frequency-domain BSS problem for convolutive acoustic mixtures in several adverse conditions. Through the real-world experiments, we show the superiority of the presented algorithm among other popular algorithms in the literature in terms of the performance and complexity computation. Moreover, we compared several permutation solvers in terms of computation complexity and performance to provide the RobustICA-based algorithm with an efficient frequency-dependent permutation scheme. Finally, we studied the effect of several parameters on the separation performance of the presented algorithm. We also presented the effect of the type of the window on the separation performance and we also showed that the performance improves at a certain range of overlapping between the signals. Lastly, in this paper, we showed the performance of a system that can work efficiently with around 0.5–10 seconds of input data, which is close to the real-time implementation. Accordingly, the presented algorithm is optimized to be suitable for the real-time operation. As a result, it is suitable for a large number of applications to ensure the real-time implementation.

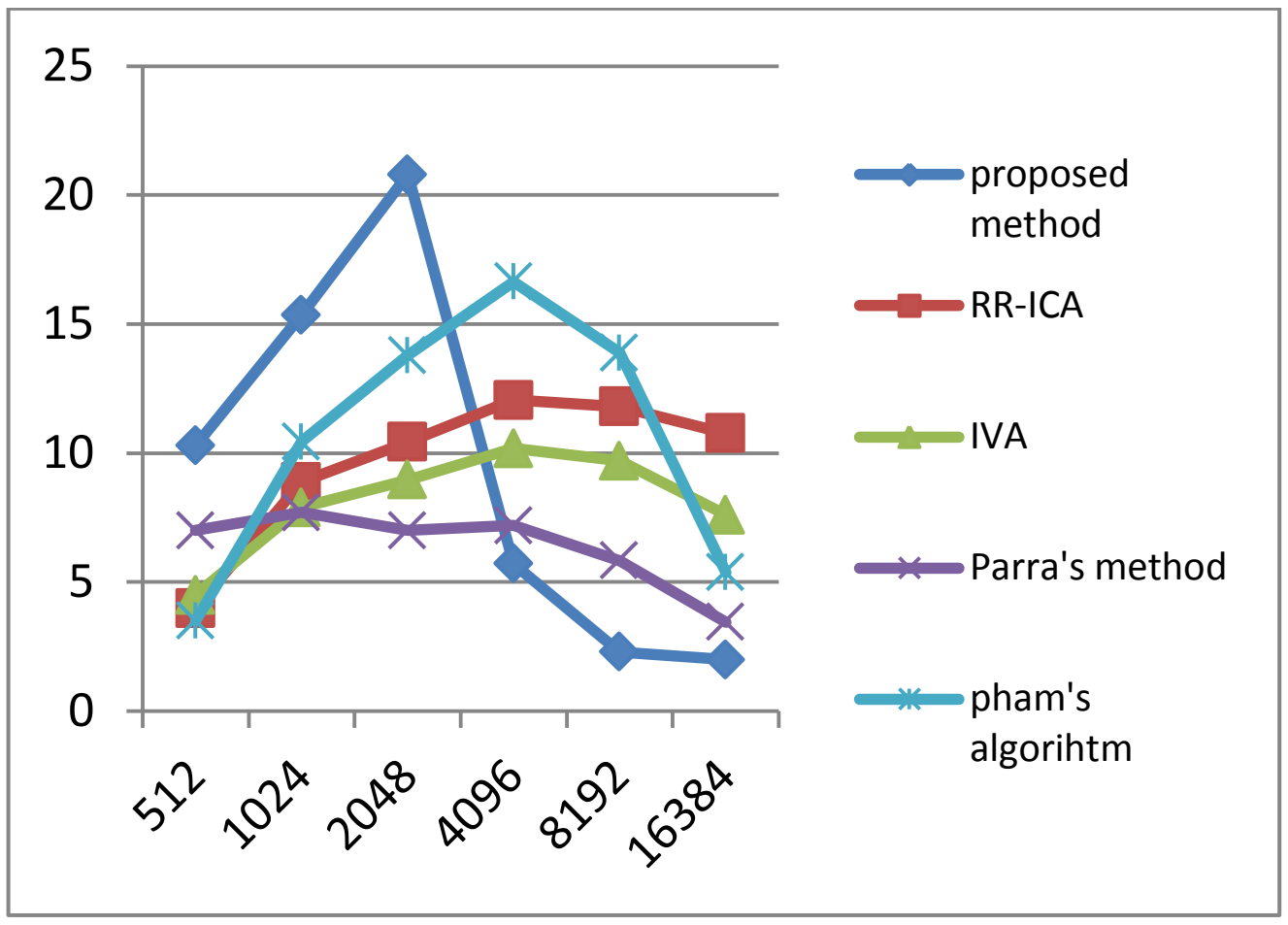

Fig. 11. Impact of FFT length, 2-by-2 case, Results obtained in the Test2 experiments.